\documentclass[runningheads]{llncs}
\usepackage[T1]{fontenc}
\usepackage{graphicx}
\usepackage{fontawesome} 
\usepackage{amsmath, amsfonts, amssymb} 
\usepackage{xcolor}
\usepackage{tabularx}

\usepackage{makecell}
\usepackage{booktabs}
\usepackage[table]{xcolor}
\usepackage{verbatim}
\usepackage{array}
\usepackage{pifont}
\usepackage{colortbl}
\usepackage{subcaption}
\usepackage{bm}
\usepackage{hyperref}

\newcommand{\cmark}{\ding{51}}
\newcommand{\xmark}{\ding{55}}

\newcommand{\flame}{\raisebox{-0.2ex}{\includegraphics[height=1em]{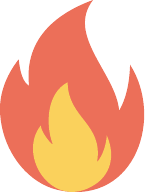}}}
\newcommand{\snow}{\raisebox{-0.2ex}{\includegraphics[height=1em]{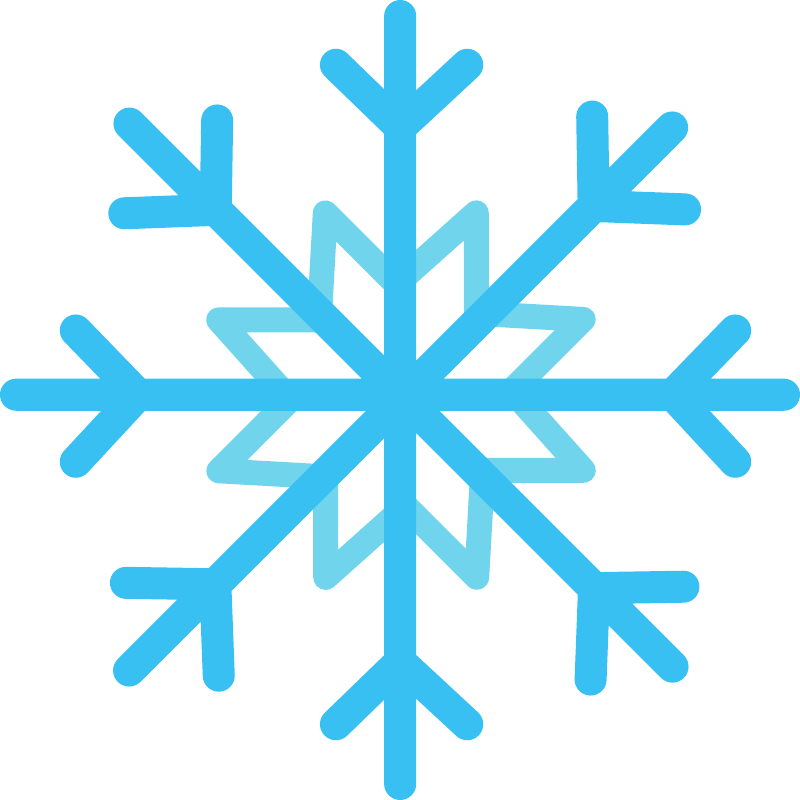}}}
\newcommand{\secbest}[1]{\smash[b]{\underline{#1}}}

\newcommand{\rot}[1]{\rotatebox[origin=b]{90}{\makebox[2cm][l]{\raggedright\strut #1}}}
\newcolumntype{C}[1]{>{\centering\arraybackslash}p{#1}}
\newcommand{\avgstd}[2]{\raisebox{-0.575em}{\shortstack[c]{#1\\[-1pt]{\tiny$\pm$\,#2}}}}
\newcolumntype{G}[1]{>{\columncolor{gray!7}}C{#1}}
\definecolor{lightgray}{gray}{0.9}

\begin{document}
\title{
HAC: Parameter-Efficient Hyperbolic Adaptation of CLIP for Zero-Shot VQA}
\titlerunning{Parameter-Efficient Hyperbolic Adaptation of CLIP for Zero-Shot VQA}
%
\author{Francesco~Dibitonto\inst{1,2}\orcidID{0009-0006-7686-8860} \and
Cigdem~Beyan\inst{1}\orcidID{0000-0002-9583-0087} \and
Vittorio~Murino\inst{1,3}\orcidID{0000-0002-8645-2328}}

\authorrunning{F. Dibitonto et al.}
%
\institute{Department of Computer Science, University of Verona, Verona, Italy \and
EVS - Embedded Vision Systems Srl, Verona, Italy \and 
AI for Good (AIGO), Istituto Italiano di Tecnologia, Genova, Italy
\\
\email{\{francesco.dibitonto, cigdem.beyan, vittorio.murino\}@univr.it}}
\maketitle              
%
\begin{abstract}

Recent advances in representation learning have shown that hyperbolic geometry can offer a more expressive alternative to the Euclidean embeddings used in CLIP models, capturing hierarchical structures and leading to better-organized representations. However, current hyperbolic CLIP variants are trained entirely from scratch, which is computationally expensive and resource-intensive. In this work, we propose HAC (Hyperbolic Adaptation of CLIP), a parameter-efficient framework that enables pretrained CLIP models to transition into hyperbolic space via lightweight fine-tuning. 
We apply HAC to Visual Question Answering (VQA), where models must interpret visual elements and align them with textual queries. Notably, HAC’s training is performed on a dataset with no overlap with any VQA benchmark, resulting in a strict zero-shot evaluation paradigm that underscores HAC’s task-agnostic adaptability. 
We evaluate HAC across a diverse suite of VQA benchmarks spanning General, Reasoning, and OCR categories. Both HAC-S (small) and HAC-B (medium) consistently surpass Euclidean baselines and prior hyperbolic approaches, with HAC-B delivering up to a +1.9 point average improvement over CLIP-B on reasoning-intensive tasks. Our code is available at \href{https://github.com/fdibiton/HAC}{https://github.com/fdibiton/HAC}
%

\keywords{Visual Question Answering  \and Hyperbolic Geometry \and CLIP \and Representation Learning \and Adapters \and Parameter-Efficient Fine-Tuning}
\end{abstract}
\section{Introduction}



Contrastive Language-Image Pretraining (CLIP)~\cite{DBLP:conf/icml/RadfordKHRGASAM21} is a vision–language model that jointly embeds images and text in a shared space via contrastive learning. Trained on large collections of image–caption pairs, CLIP learns to align matched image–text representations while separating mismatched ones, enabling strong zero-shot generalization across downstream tasks. Beyond image classification and retrieval, recent work has extended CLIP’s zero-shot capabilities to Visual Question Answering (VQA)~\cite{DBLP:conf/iccv/AntolALMBZP15} by reformulating questions as descriptive prompts~\cite{DBLP:conf/acl/0002000W22} or decomposing them into reasoning steps handled by external pretrained models~\cite{DBLP:conf/acl/Cao023}.


Despite the success of CLIP and its zero-shot VQA extensions, its embedding space remains Euclidean. Euclidean spaces treat distances uniformly and exhibit only polynomial volume growth, making them poorly suited for representing hierarchical or tree-structured relationships. Hyperbolic geometry, with its negative curvature and exponential expansion, provides a natural manifold for organizing concepts according to their level of abstraction~\cite{DBLP:conf/nips/NickelK17,DBLP:conf/icml/GaneaBH18}.
Previous work~\cite{DBLP:conf/icml/DesaiNR0V23,DBLP:conf/iclr/PalSMFGM25} has shown that incorporating hyperbolic geometry into multimodal embedding spaces yields measurable gains in CLIP zero-shot classification and retrieval. Motivated by these findings, we extend hyperbolic representations to VQA, leveraging their capacity to encode hierarchical and relational structure among objects and scenes, an ability we argue is crucial for understanding and answering visual queries.



Existing hyperbolic CLIP variants, however, are trained entirely from scratch \cite{DBLP:conf/icml/DesaiNR0V23,DBLP:conf/iclr/PalSMFGM25}, a process that demands substantial computational resources and limits their practical deployment. 
To overcome this barrier, we introduce \textbf{HAC} (\textbf{H}yperbolic \textbf{A}daptation of \textbf{C}LIP), a lightweight and resource-efficient framework that lifts pretrained Euclidean CLIP models into hyperbolic space. HAC employs parameter-efficient adaptation modules that reshape the embedding geometry while keeping CLIP’s large pretrained backbones frozen, thereby requiring minimal computation. These modules include adapters~\cite{DBLP:conf/icml/HoulsbyGJMLGAG19,DBLP:conf/iclr/HeZMBN22}, low-rank updates~\cite{DBLP:conf/iclr/HuSWALWWC22}, and component-specific edits such as bias~\cite{DBLP:conf/acl/ZakenGR22} and LayerNorm tuning~\cite{DBLP:conf/iclr/ZhaoT0MX24}. Through these localized modifications, CLIP’s Euclidean features are reorganized within the Lorentz hyperbolic model~\cite{Ratcliffe2006FoundationsOH}, enabling more expressive and hierarchy-aware representations.
Building on this adapted geometry, HAC performs VQA in a purely zero-shot manner by comparing image and question–answer embeddings via hyperbolic distance, without prompt reformulation or auxiliary modules. HAC consistently improves over Euclidean CLIP: across six VQA benchmarks, our HAC-S (small) model matches or surpasses the CLIP-S baseline on all tasks, and our HAC-B (medium) model 
outperforms CLIP-B on reasoning-intensive tasks by +1.9 average points.

\noindent The main contributions of our work are the following. 
\vspace{-1.3em}
\begin{itemize}
    \item \textbf{First hyperbolic adaptation of CLIP:}
    We introduce \textbf{HAC}, the first Euclidean-to-hyperbolic transition framework that lifts pretrained CLIP models into hyperbolic space \emph{without} retraining them from scratch.
    \item \textbf{A systematic study of parameter-efficient 
    adaptation in hyperbolic geometry:}
    We conduct the first comprehensive evaluation of parameter-efficient adaptation strategies for hyperbolic modeling, comparing adapters, low-rank updates, and component-level tuning to identify the most effective pathway for transitioning CLIP into non-Euclidean geometry.
    \item \textbf{State-of-the-art zero-shot VQA performance under hyperbolic geometry:}
    HAC achieves consistent improvements over Euclidean CLIP and prior hyperbolic variants across General, Reasoning, and OCR VQA benchmarks, while using far fewer trainable parameters.
\end{itemize}

\section{Related Work}
\noindent \textbf{Zero-shot VQA with CLIP.}
Early efforts to use CLIP for VQA showed limitations in directly applying it to question answering. Shen et al.~\cite{DBLP:conf/iclr/ShenLTBRCYK22} tested OpenAI's CLIP models~\cite{DBLP:conf/icml/RadfordKHRGASAM21} on zero-shot VQA by constructing the prompt template ``question: [question text] answer: [answer text]'' and observed near-chance performance, suggesting the need for additional pretraining or fine-tuning. Building on this observation, Song et al.~\cite{DBLP:conf/acl/0002000W22} reformulate each question as a masked statement and fill the mask with candidate answers, improving zero-shot results. Cao and Jiang~\cite{DBLP:conf/acl/Cao023} propose a composite framework in which additional pretrained modules support CLIP's localization and spatial reasoning. In contrast, our approach directly queries the model without prompt reformulation or auxiliary modules, achieving competitive results by leveraging object-scene hierarchies. \\

\noindent \textbf{Hyperbolic CLIP models.}
Recent works 
embed CLIP in the hyperbolic space, leveraging its exponential geometry to encode the hierarchical structure of visual and textual concepts.
MERU~\cite{DBLP:conf/icml/DesaiNR0V23} reformulates CLIP to hyperbolic geometry by mapping image-text embeddings from Euclidean to Lorentz space. The model is trained from scratch using a contrastive loss based on negative Lorentzian distance, together with an entailment loss that imposes a partial order between text and image representations. This formulation enables MERU to encode semantic hierarchies and yields improvements in zero-shot image classification and retrieval over equivalently trained Euclidean CLIP baselines.
HyCoCLIP~\cite{DBLP:conf/iclr/PalSMFGM25} extends hyperbolic contrastive learning by introducing a compositional entailment objective. It models a compositional hierarchy in which objects, treated as more generic components, are embedded closer to the origin of hyperbolic space, while scenes or image-level representations lie farther from the origin as more specific instantiations. HyCoCLIP is trained on the GRIT dataset~\cite{peng2024grounding}, which provides grounded image–text pairs and localized object annotations that support learning such structure. The resulting representations yield further improvements on zero-shot benchmarks compared to MERU~\cite{DBLP:conf/icml/DesaiNR0V23}.
In our work, we follow HyCoCLIP’s compositional design and use GRIT to instantiate object–scene hierarchies. However, unlike MERU~\cite{DBLP:conf/icml/DesaiNR0V23} and HyCoCLIP~\cite{DBLP:conf/iclr/PalSMFGM25}, our method is substantially more practical: (a) it adapts pretrained CLIP models rather than training from scratch, (b) requires far fewer trainable parameters, and (c) introduces no auxiliary modules or prompt reformulation. This lightweight design enables efficient hyperbolic modeling while preserving CLIP’s generalization capabilities, allowing HAC to transfer zero-shot to VQA without using any VQA supervision. \\

\noindent \textbf{Parameter-Efficient Fine-Tuning.}
PEFT~\cite{DBLP:conf/icml/HoulsbyGJMLGAG19} has emerged as a widely used strategy for adapting large pretrained models, such as Transformers, across various domains while updating only a small subset of parameters. BitFit~\cite{DBLP:conf/acl/ZakenGR22} tunes only bias terms, while layernorm tuning~\cite{DBLP:conf/iclr/ZhaoT0MX24} updates normalization parameters. Adapter-based methods add lightweight bottleneck modules into Transformer layers, either sequentially~\cite{DBLP:conf/icml/HoulsbyGJMLGAG19,madClip} or in parallel~\cite{DBLP:conf/iclr/HeZMBN22,madPOT}, enabling task-specific adaptation with the base model kept frozen. LoRA~\cite{DBLP:conf/iclr/HuSWALWWC22} instead inserts low-rank matrices that approximate full-weight updates more efficiently. Although PEFT has been successfully applied in numerous settings, these techniques have not been explored for hyperbolic modeling or for geometrically transforming pretrained CLIP models. In this work, we systematically evaluate PEFT strategies to identify the most effective approach for Euclidean-to-hyperbolic adaptation.

\section{Preliminaries}
\vspace{-0.5em}
Hyperbolic space provides a mathematical setting for representing hierarchical and tree-like structures. Unlike Euclidean space, which has zero curvature, hyperbolic manifolds have negative curvature, causing distances and volumes to grow exponentially with radius~\cite{Ratcliffe2006FoundationsOH,Lee2019IntroductionTR}. This exponential expansion makes hyperbolic space well suited for encoding hierarchical relationships: general concepts occupy regions near the origin, while increasingly specific instances appear deeper in the manifold, mirroring the structure of tree-like semantic hierarchies~\cite{DBLP:conf/nips/NickelK17}. \\

\noindent \textbf{The Lorentz Model.} Also known as the hyperboloid or Minkowski model~\cite{Ratcliffe2006FoundationsOH}, is a common formulation of hyperbolic space. It embeds an \textit{n}-dimensional hyperbolic manifold $\mathbb{L}^n$ into an (\textit{n}+1)-dimensional Minkowski space $\mathbb{R}^{n,1}$, defined as: $\mathbb{L}^n = \{\, x \in \mathbb{R}^{n,1} : \langle x, x \rangle_L = -\tfrac{1}{\kappa},\; x_0 > 0 \,\}$.
Here, $\kappa > 0$ is the curvature magnitude. The Lorentzian inner product is: $\langle x, y \rangle_L = -x_0 y_0 + \sum_{i=1}^{n} x_i y_i$,
where $x_0$ represents a ``time'' dimension, while the remaining coordinates $(x_1,\allowbreak x_2,\allowbreak \ldots,\allowbreak x_n)$ correspond to spatial dimensions.
The model defines a two-sheeted hyperboloid, where the upper sheet ($x_0$ > 0) represents points in the hyperbolic space.
From a geometric perspective, the time component controls the vertical direction of the hyperboloid, while the spatial components spread points in the surrounding radial directions. This formulation is widely used in learning settings, as it allows numerically stable computation of hyperbolic distances and mappings.
\\\\
\noindent \textbf{Tangent Space and Exponential Mapping.} At each point $p \in \mathbb{L}^n$, the tangent space $T_p\mathbb{L}^n$ locally approximates the manifold with Euclidean geometry and is defined as: $T_p\mathbb{L}^n = \{\, v \in \mathbb{R}^{n,1} : \langle v, p \rangle_L = 0 \,\}$.
The exponential map can connect the tangent space to the manifold: it projects a tangent vector $v$ onto the hyperboloid and is defined as:
\begin{equation}
\exp_p^{\kappa}(v) = 
\cosh\!\left(\sqrt{\kappa}\,\|v\|_L\right)p + 
\frac{\sinh\!\left(\sqrt{\kappa}\,\|v\|_L\right)}{\sqrt{\kappa}\,\|v\|_L}\,v,
\label{eq:exp_map}
\end{equation}
where $\|v\|_L = \sqrt{|\langle v, v \rangle_L|}$ denotes the Lorentzian norm. \\

\noindent \textbf{Geodesic Distance.} The distance between two points $x, y \in \mathbb{L}^n$ is measured by the Lorentzian geodesic distance, and it is defined as: 
\begin{equation}
d_L(x, y) = \sqrt{\frac{1}{\kappa}} \, \cosh^{-1}\!\left(-\kappa \, \langle x, y \rangle_L \right),
\label{eq:lorentz_distance}
\end{equation}
which corresponds to the length of the shortest path on the hyperboloid surface. \\

\noindent \textbf{Entailment cones.} 
Hyperbolic entailment cones provide a mechanism for expressing hierarchical relations in hyperbolic geometry. Following the formulation of Ganea et al.~\cite{DBLP:conf/icml/GaneaBH18}, each point $x$ on the manifold is associated with a convex cone that expands outward from $x$ and encodes the set of points entailed by it. A cone at $x$ is parameterized by a half-aperture $\psi(x)$, which determines how wide the cone opens. Formally, for tangent vectors $v \in T_x\mathbb{L}^n$, the cone is defined as $S_x = \exp_x\{\, v \in T_x\mathbb{L}^n : \angle(v, \bar{x}) \le \psi(x) \,\}$,
where $\bar{x}$ denotes the cone axis, i.e., the normalized tangent direction from $x$ toward the origin. The angular constraint $\angle(v, \bar{x}) \le \psi(x)$ restricts the allowable tangent directions to those lying within $\psi(x)$ 
from the axis. Applying the exponential map $\exp_x$ then projects this tangent-space cone onto the hyperbolic manifold, yielding the hyperbolic entailment cone $S_x$. This construction induces a hierarchy where more general concepts have wider cones that subsume those of more specific concepts.

\section{Our Method}
\vspace{-0.5em}
\subsection{Problem Formulation}
\label{subsec:prob_form}
Given a dataset $\mathcal{D} = \{(I_i, T_i)\}_{i=1}^{N}$ of $N$ image-text pairs, a CLIP model~\cite{DBLP:conf/icml/RadfordKHRGASAM21} learns to align visual and textual representations in a shared embedding space. Each image $I_i$ is associated with a textual description $T_i$, and both modalities are processed by separate encoders to produce representations in a joint space.

In our setting, we follow~\cite{DBLP:conf/iclr/PalSMFGM25} and consider image-text pairs that provide both global scene-level information and localized object annotations. This structure allows the model to capture hierarchical relations between objects and their surrounding contexts. Formally, each sample can be expressed as $(I, T, I_{\text{box}}, T_{\text{box}})$, where $(I, T)$ denotes the full image and caption pair, and $(I_{\text{box}}, T_{\text{box}})$ represent corresponding localized object regions and their textual descriptions. \\

\noindent \textbf{Model Architecture.}
Our approach follows the dual-encoder formulation of CLIP~\cite{DBLP:conf/icml/RadfordKHRGASAM21}, consisting of an image encoder $f_I(\cdot)$ and a text encoder $f_T(\cdot)$. The two encoders produce modality-specific embeddings, which are linearly projected to a shared latent space: $v_I = f_I(I)$ and $v_T = f_T(T)$.
In the Euclidean formulation, these embeddings are L2-normalized and compared by cosine similarity.
In our method, they are instead lifted directly to hyperbolic space, as detailed below. 

\subsection{Hyperbolic Setup}
\label{subsec:hyp_setup}
To encode hierarchical relationships, we embed image and text representations in the Lorentz model of hyperbolic geometry.
Following~\cite{DBLP:conf/icml/DesaiNR0V23}, we adopt the coordinate order $[x_{\text{space}}, x_{\text{time}}]$ in the Lorentz model, and we augment each Euclidean embedding $v_{euc} \in \mathbb{R}^{n}$ with a zero time component, forming: $\,v = [v_{\text{euc}}, 0] \in \mathbb{R}^{n+1}\,$.
This parameterization ensures that $v$ lies in the tangent space at the origin of the hyperboloid, allowing it to be mapped onto the manifold using the exponential map from Eq.~\ref{eq:exp_map}.
The exponential map then reduces to the simplified form presented in~\cite{DBLP:conf/icml/DesaiNR0V23}:
\begin{equation}
x_{\text{space}} = 
\frac{\sinh\!\left(\sqrt{\kappa}\,\|v_{\text{euc}}\|\right)}
{\sqrt{\kappa}\,\|v_{\text{euc}}\|}\,v_{\text{euc}},
\label{eq:exp_map_simplified}
\end{equation}
which maps Euclidean embeddings onto the hyperboloid $\mathbb{L}^n \subset \mathbb{R}^{n,1}$. The full derivations and simplifications can be found in~\cite{DBLP:conf/icml/DesaiNR0V23}. 
To avoid numerical overflow, we scale the Euclidean embeddings $v_{\text{euc}}$ using learnable projection scalars $\alpha_{\text{img}}$ and $\alpha_{\text{txt}}$ before applying the exponential map~\cite{DBLP:conf/icml/DesaiNR0V23}.
Both scalars are initialized to $1/\sqrt{n}$, where $n$ is the embedding dimension, and are optimized in log-space.

\subsection{Zero-shot VQA Formulation}
\label{subsec:vqa_setup}

We formulate zero-shot VQA as a multiple-choice matching problem and define a VQA evaluation dataset as follows: $\mathcal{D}_{\mathrm{VQA}} = \bigl\{\, (I_k,\, q_k,\, \{a_{k,1}, \ldots, a_{k,n}\},\, a_k^{\star}) \,\bigr\}_{k=1}^{M}$,
where each example consists of an image $I_k$, a question $q_k$, a set of $n$ candidate answers, and the ground-truth answer $a_k^{\star}$.
This evaluation format differs from the training data described in Sec.~\ref{subsec:prob_form}, which provides image–text pairs and object annotations for learning hierarchical representations.

For each question $q_k$ and its candidates $\{a_{k,1}, \ldots, a_{k,n}\}$, we form the list of textual queries: $\mathcal{P}(q_k) = \{\, q_k \texttt{[SEP]} a_{k,1},\, \ldots,\, q_k \texttt{[SEP]} a_{k,n} \,\}$,
where \texttt{[SEP]} is a simple whitespace.
Each element in $\mathcal{P}(q)$ is processed by the text encoder and lifted onto the hyperboloid using Eq.~\ref{eq:exp_map_simplified}, yielding the hyperbolic query embeddings $\{\, f_T(q\texttt{[SEP]}a_j) \,\}_{j=1}^{n}$.
The image $I_k$ is encoded using the same exponential mapping, producing a hyperbolic image embedding $f_I(I_k)$. 
The predicted answer is obtained by selecting the hyperbolic query embedding with the smallest Lorentzian geodesic distance to $f_I(I_k)$ (Eq.~\ref{eq:lorentz_distance}). In practice, this is implemented by maximizing the negative Lorentzian distance for convenience:
\vspace{-0.4em}
\begin{equation}
\
\hat{a}
=
\arg\max_{a_j}
\; -\, d_L\!\left( f_I(I),\, f_T(q\texttt{[SEP]}a_j) \right),
\
\vspace{-0.8em}
\end{equation}
VQA zero-shot accuracy is then computed as the fraction of correctly predicted question-image pairs in the dataset: $\mathrm{Acc}_{\mathrm{VQA}}
= \frac{1}{M} \sum_{k=1}^{M} \mathbf{1}\!\left[\, \hat{a}_k = a_k^{\star} \,\right]$,
where $\mathbf{1}[\cdot]$ is the indicator function, assigning $1$ to correct and $0$ to incorrect predictions.

\subsection{Hyperbolic Adaptation via Parameter-Efficient Fine-Tuning}
\label{subsec:peft}
\begin{figure}[t!]
    \centering
    \includegraphics[width=1.0\linewidth]{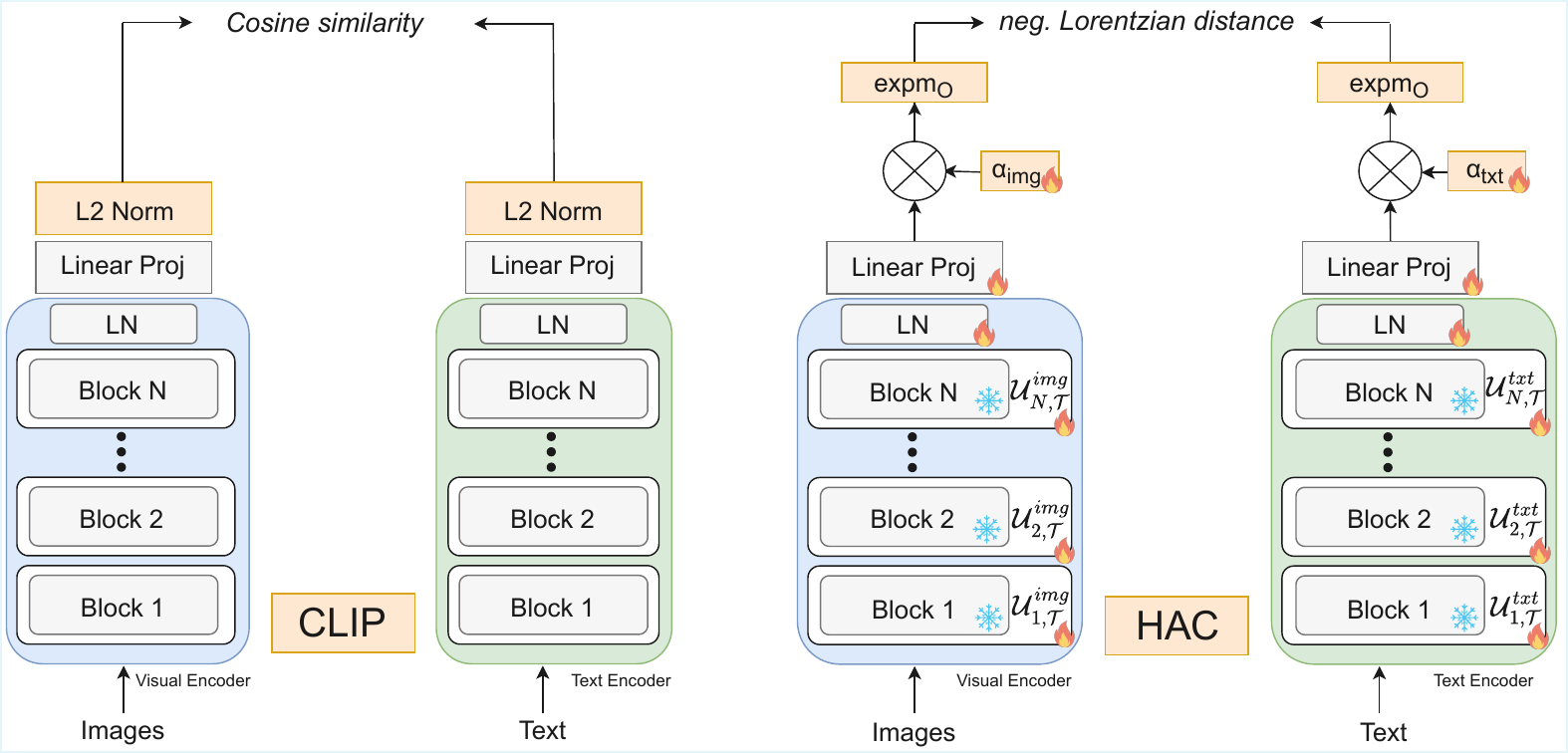}
    \caption{\textbf{Overview of our HAC} enabling geometric adaptation of pretrained CLIP models through parameter-efficient fine-tuning. HAC updates a limited number of Adaptation parameters that are added by wrapping each selected block $\ell$ with an Adaptation module $\mathcal{U}_{\ell,\mathcal{T}}$, where $\mathcal{T}$ is a lightweight transformation of the block. HAC also fully trains the final LayerNorm, Linear Projection layers, and Projection $\alpha$ scalars. \flame\ denotes trainable and \snow\ denotes frozen parameters.}
    \label{fig:hac_method}
    \vspace{-1em}
\end{figure}

To enable geometric adaptation of pretrained CLIP models, we define \textbf{HAC} (\textbf{H}yperbolic \textbf{A}daptation of \textbf{C}LIP), a generic PEFT~\cite{DBLP:conf/icml/HoulsbyGJMLGAG19} framework. The idea behind our HAC is to frame the Euclidean to hyperbolic transition as a fine-tuning objective focused on reshaping the embedding geometry, without altering representational knowledge already learned by Euclidean CLIP. We assume that this geometric reshaping can be realized through localized parameter-efficient updates rather than full retraining, as the semantics captured by Euclidean CLIP remain valid and only the geometric organization of the embedding space requires adjustment.
An overview of our HAC framework is presented in Fig.~\ref{fig:hac_method}.

Given a Transformer-based CLIP architecture consisting of an image encoder $f_I$ and a text encoder $f_T$, our goal is to introduce only a small number of trainable parameters while leaving the pretrained backbones of $f_I$ and $f_T$ unchanged. 
\\

\begin{figure}[t!]
    \centering
    \includegraphics[width=1.0\linewidth]{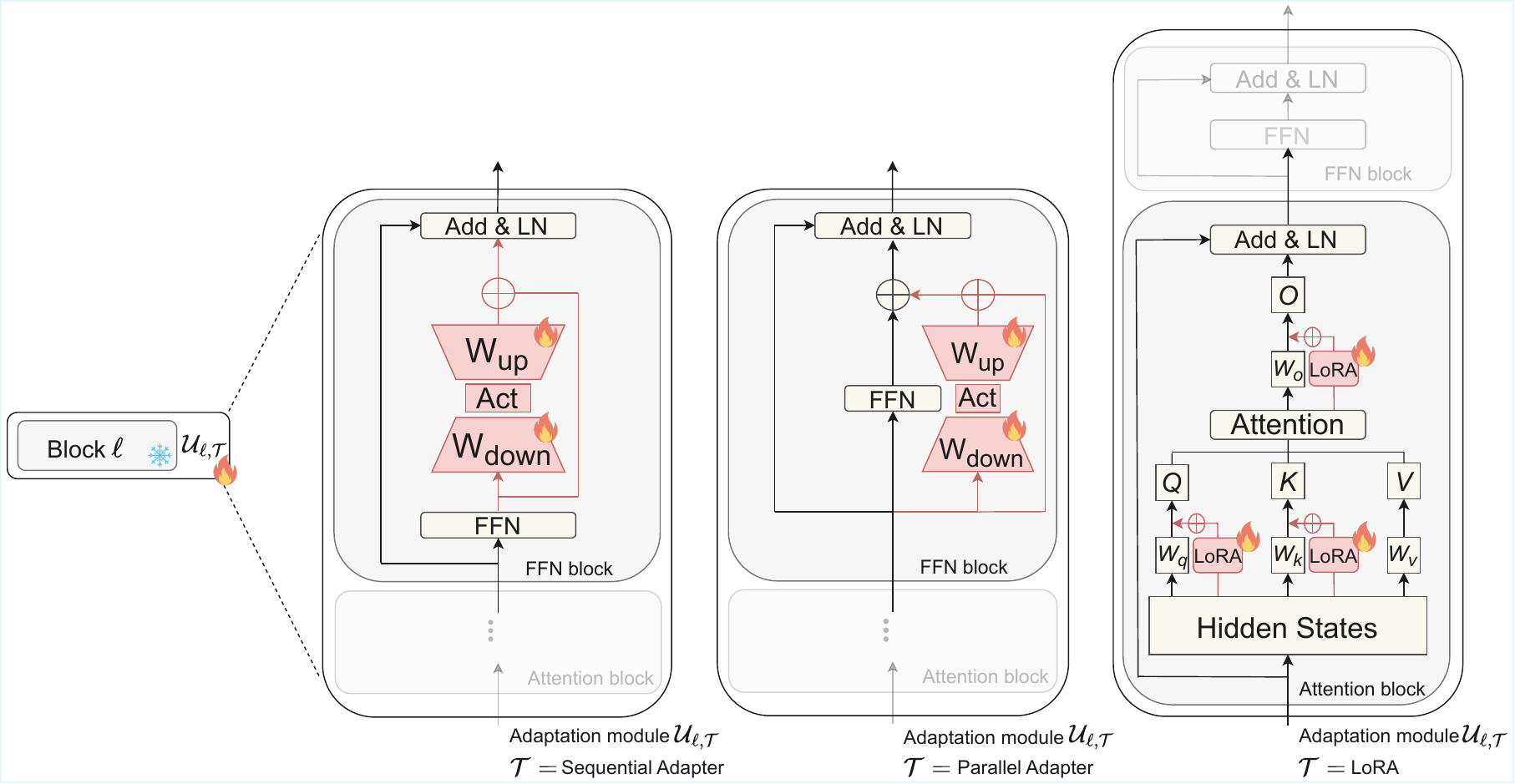}
    \caption{\textbf{HAC Adapted Transformer Blocks}: For a selected block $\ell$, the HAC Adaptation module $\mathcal{U}_{\ell,\mathcal{T}}$ can implement any lightweight transformation, such as a sequential (left), a residual (middle) or a low-rank transformation (right) to selected block's submodules. \flame\ denotes the only trainable parameters.}
    \label{fig:hac_modules}
    \vspace{-1em}
\end{figure}

\noindent \textbf{Adapted Transformer Blocks.} Let $\mathcal{B}_\ell(\theta_\ell), 
\ell = 1,\dots,L$, be the $\ell$-th pretrained Transformer block, 
where $\theta_\ell$ denotes its pretrained parameters and remain unchanged during training. We introduce a generic modification of the block through an adaptation module $\mathcal{U}_{\ell,\mathcal{T}}(\cdot,\phi_\ell)$, where $\phi_\ell$ are the trainable parameters of the module. 
$\mathcal{U}_{\ell,\mathcal{T}}$ acts as a wrapper of the pretrained block $\mathcal{B}_\ell$, and it operates by applying a lightweight transformation $\mathcal{T}$ to selected subcomponents of $\mathcal{B}_\ell$, for example, a sequential transformation~\cite{DBLP:conf/icml/HoulsbyGJMLGAG19}, a residual transformation~\cite{DBLP:conf/iclr/HeZMBN22}, or low-rank update to individual projection matrices~\cite{DBLP:conf/iclr/HuSWALWWC22}, as shown in Fig.~\ref{fig:hac_modules}.
Let $\mathcal{S} \subseteq {1,\dots,L}$ denote the subset of layers that are adapted. For each $\ell \in \mathcal{S}$, we define the adapted block as follows: $\tilde{\mathcal{B}}_\ell(\theta_\ell,\phi_\ell)
= \mathcal{U}_{\ell,\mathcal{T}}(\mathcal{B}_\ell(\theta_\ell),\phi_\ell)$ (see Fig. 2).
For layers that are not adapted $\ell \notin \mathcal{S}$, we keep the pretrained block unchanged. \\

\noindent \textbf{Projection Heads.}
While CLIP's internal layers are adapted using para\-meter-eff\-icient modules, we fully retrain the final projection layers that map encoder outputs into hyperbolic space. These layers serve as the interface between CLIP’s representations and the hyperbolic manifold, and full training provides the flexibility required for this geometric transition. We also retrain the final LayerNorm~\cite{ba2016layer} of each encoder, which precedes the projection heads. Because this LayerNorm is not bypassed by a skip connection~\cite{DBLP:conf/cvpr/HeZRS16}, leaving it frozen would impose a fixed gating effect on the encoder’s outputs.
These projection heads and final LayerNorms are the only components fully trained in our framework. 

\subsection{Training Objective}
\label{subsec:train_obj}
To train our HAC, we incorporate the hyperbolic objectives introduced in HyCoCLIP~\cite{DBLP:conf/iclr/PalSMFGM25} and adapt them to our CLIP-to-hyperbolic transition framework.
We use the hierarchical Compositional Contrastive loss $\mathcal{L}_{\mathrm{hCC}}$ to jointly align full images with captions and object regions with their textual descriptions, and the hierarchical Compositional Entailment loss $\mathcal{L}_{\mathrm{hCE}}$ to learn object–scene hierarchies within our adapted hyperbolic space. The total training loss is $\mathcal{L} = \mathcal{L}_{\mathrm{hCC}} + \lambda\,\mathcal{L}_{\mathrm{hCE}}$
where $\lambda$ controls the contribution of the entailment term.
Here, $\mathcal{L}_{\mathrm{hCC}}$ penalizes non-matching image-text correspondences at both scene and object level, analogous to CLIP~\cite{DBLP:conf/icml/RadfordKHRGASAM21} but using hyperbolic distances. $\mathcal{L}_{\mathrm{hCE}}$ penalizes deviations from the intended hierarchy, where objects encode general concepts near the origin and scenes encode specific instantiations of those objects at larger radii. It does so by measuring how far a scene embedding lies outside its parent object's entailment cone, assigning a penalty proportional to this deviation.

This compositional objective naturally aligns with the requirements of VQA, where reasoning typically depends on understanding objects and their surrounding context. By construction, the objective induces a representational separation between objects and full scenes: objects are encouraged to acquire their own dedicated geometric space, distinct from the space used to encode scene-level information. This encourages the model to develop object-specific structures that are not entangled with global context. Consequently, when a question refers to an object appearing in a new or unseen setting, the model can rely on a stable, context-independent object representation to interpret the query.

\vspace{-1em}

\section{Experiments}

The primary goal of our experiments is to assess whether transitioning pretrained CLIP encoders to the hyperbolic space leads to improved zero-shot VQA performance compared to maintaining their original Euclidean geometry. To this end, we run a set of experiments to determine which PEFT method is more suitable for this transition. We also compare the selected HAC configurations against Euclidean CLIP baselines and state-of-the-art (SOTA) fully hyperbolic models.

\subsection{Implementation Details}
\noindent \textbf{Baselines.} 
All experiments are conducted using the GRIT dataset~\cite{peng2024grounding}, which provides image–caption pairs together with grounded object annotations. We compare our method against Euclidean CLIP baselines that were trained on GRIT~\cite{DBLP:conf/iclr/PalSMFGM25}.
Due to dataset decay, a portion of the GRIT samples is no longer accessible online. After filtering missing entries, our collected version contains approximately 13.7M usable samples—about 66\% of the original 20.5M examples reported in the release.
Whereas the Euclidean baselines were trained on the full dataset, all HAC models are trained on this 13.7M subset. Consequently, HAC’s performance reflects training under a more \textbf{data-limited regime}. \\
\vspace{-0.5em}

\noindent \textbf{SOTA Hyperbolic Models.}
We also compare our method against fully hyperbolic CLIP variants. Specifically, we include MERU~\cite{DBLP:conf/icml/DesaiNR0V23}, which is trained from scratch on 12M Redcaps samples~\cite{DBLP:conf/nips/DesaiKA021}, and HyCoCLIP~\cite{DBLP:conf/iclr/PalSMFGM25}, which is trained end-to-end on the complete 20.5M GRIT dataset~\cite{peng2024grounding}. 
Note that HyCoCLIP~\cite{DBLP:conf/iclr/PalSMFGM25} is trained with full-parameter updates and significantly larger compute, while HAC relies on parameter-efficient adaptation of pretrained CLIP. As such, HyCoCLIP represents a stronger upper-bound reference trained under more favorable conditions, rather than a directly comparable counterpart. \\
\vspace{-0.5em}

\noindent \textbf{HAC Models.} We consider two HAC configurations that differ in image encoder capacity while sharing the same 12-layer, 512-dimensional text Transformer. HAC-S (small) uses a ViT-S encoder, whereas HAC-B (medium) uses the larger ViT-B.
Both models are initialized from their corresponding Euclidean CLIP variants trained on GRIT~\cite{DBLP:conf/iclr/PalSMFGM25}: CLIP-S for HAC-S and CLIP-B for HAC-B. We set the coefficient $\lambda$ for the entailment loss to 0.1 as in~\cite{DBLP:conf/iclr/PalSMFGM25}, and initialize the learnable projection scalars $\alpha_{\text{img}}$, $\alpha_{\text{txt}}$, and the curvature parameter $k$ following~\cite{DBLP:conf/icml/DesaiNR0V23}.
Finally, for both HAC configurations, we randomly re-initialize CLIP’s projection heads and the final LayerNorm~\cite{ba2016layer} of each encoder. \\
\vspace{-0.5em}

\noindent \textbf{PEFT Methods.} We examine \emph{(i) bias tuning (BitFit)}~\cite{DBLP:conf/acl/ZakenGR22}, \emph{(ii) LayerNorm tuning}~\cite{DBLP:conf/iclr/ZhaoT0MX24}, \emph{(iii) sequential adapters}~\cite{DBLP:conf/icml/HoulsbyGJMLGAG19}, \emph{(iv) parallel adapters}~\cite{DBLP:conf/iclr/HeZMBN22} and \emph{(v) LoRA}~\cite{DBLP:conf/iclr/HuSWALWWC22}. We implement sequential~\cite{DBLP:conf/icml/HoulsbyGJMLGAG19} and parallel adapters~\cite{DBLP:conf/iclr/HeZMBN22} using the Adapters library~\cite{poth-etal-2023-adapters}, adopting the default hyperparameters and setting the bottleneck dimension to 16.
For LoRA~\cite{DBLP:conf/iclr/HuSWALWWC22}, we set $r=\alpha=8$, we use rank stabilization~\cite{Kalajdzievski2023ARS} and apply low-rank updates only to query and value projection matrices.
We apply all PEFT methods to all 12 layers of both encoders, with the exception of LoRA, which we apply to every layer except the first to avoid out-of-memory issues. For the larger HAC-B configuration, we further refine hyperparameters. Spe\-cif\-ic\-ally, we apply LoRA to all attention submatrices (\emph{q},\emph{k},\emph{v},\emph{o}) with $r=\alpha=128$. In HAC-B, we restrict all adaptation modules to the last 4 and 8 Transformer blocks of the vision and text encoders, respectively. For all HAC configurations, we implement all PEFT strategies within the adaptation framework described in Sect.~\ref{subsec:peft}, where the adaptation module $\mathcal{U}_{\ell,\mathcal{T}}$ represents the corresponding PEFT method $\mathcal{T}$  applied to the pretrained block $\ell$. \\
\vspace{-0.5em}

\noindent \textbf{Data Augmentation.} 
We resize the shorter side of each image to 224 pixels while preserving aspect ratio, then center-crop to a 224$\times$224 square and apply a random horizontal flip. We additionally use random Gaussian blur and photometric augmentations, including contrast, color, sharpness, equalization, and gamma adjustments (see Supp. Mat. for details). For text, we apply NEFTune perturbations~\cite{DBLP:conf/iclr/JainCWKCSBKSSGG24} to all token embeddings using a noise coefficient $\alpha = 0.1$. \\
\vspace{-0.5em}

\noindent \textbf{Optimization.} We train all of our models for 30K iterations with a batch size of 768. Following~\cite{DBLP:conf/nips/DesaiKA021,DBLP:conf/iclr/PalSMFGM25}, we use the AdamW optimizer~\cite{DBLP:conf/iclr/LoshchilovH19} with $\beta_1 = 0.9$, $\beta_2 = 0.98$, and a weight decay of 0.2, which is disabled for all gains, biases, and learnable scalars. The learning rate is set to $2.5\times 10^{-4}$, using a linear warmup for the first 4K steps followed by cosine decay~\cite{DBLP:conf/iclr/LoshchilovH17} over the remaining training iterations. 

\subsection{Benchmarks and Evaluation Metrics}
We evaluate all models on six VQA benchmarks spanning three categories: \emph{General}, \emph{Reasoning}, and \emph{OCR}. Consistent with our zero-shot VQA formulation (Sec.~\ref{subsec:vqa_setup}), we select datasets in native multiple-choice format: A-OKVQA~\cite{Schwenk2022AOKVQAAB}, MMStar~\cite{NEURIPS2024_2f8ee6a3}, SEEDBench~\cite{Li2024SEEDBenchBM}, ScienceQA~\cite{lu2022learn}, RealWorldQA~\cite{realworldqa2024}, and AI2D~\cite{DBLP:conf/eccv/KembhaviSKSHF16}.
For efficient batched inference, we convert all datasets to a unified structure where each question is paired with four candidate answers. When fewer than four options are available, we duplicate an incorrect answer to complete the set.
As the evaluation metric, we use the zero-shot VQA accuracy defined in Sec.~\ref{subsec:vqa_setup}.

\subsection{Computational Complexity}
Our HAC-S model trains in under a day on a single A6000 GPU, and HAC-B likewise finishes within a day on a single A100.
In contrast, MERU requires 8$\times$V100 GPUs for nearly a full day~\cite{DBLP:conf/nips/DesaiKA021}, and HyCoCLIP requires 4$\times$A100 GPUs~\cite{DBLP:conf/iclr/PalSMFGM25}. These comparisons highlight the substantial computational efficiency of HAC.

\begin{table}[t!]
\centering
\small
\setlength{\tabcolsep}{3pt}
\renewcommand{\arraystretch}{1.05}
\caption{Zero-shot VQA results for CLIP-S variants. 
\textbf{Bold} indicate the best result, \underline{underlined} the second-best. HyCoCLIP-S is a fully trained hyperbolic model; our HAC approaches its performance while requiring far less compute.} 
\vspace{0.5em}
\resizebox{\textwidth}{!}{
\begin{tabular}{
l
C{0.90cm} C{0.90cm} 
!{\vrule width 0.6pt}
C{0.90cm} C{0.90cm} C{0.90cm} G{1.10cm} 
!{\vrule width 0.6pt}
C{0.90cm} C{0.90cm} G{1.10cm}
!{\vrule width 0.6pt}
C{0.90cm} G{1.10cm}
!{\vrule width 0.6pt}
G{1.10cm}
}

\multicolumn{1}{c}{} & \multicolumn{2}{c}{} &
\multicolumn{4}{c}{\cellcolor{gray!15}\textbf{GENERAL}} &
\multicolumn{3}{c}{\cellcolor{gray!15}\textbf{REASONING}} &
\multicolumn{2}{c}{\cellcolor{gray!15}\textbf{OCR}} &
\multicolumn{1}{c}{} \\
\cmidrule(lr){4-7}\cmidrule(lr){8-10}\cmidrule(lr){11-12}

\textbf{} &
\rot{\fontsize{8}{7.0}\selectfont train samples (M)} &
\rot{\fontsize{8}{7.0}\selectfont trainable params (M)} &
\rot{A-OKVQA} & \rot{MMStar} & \rot{SEEDBench} & \rot{Avg \tiny ± Std} &
\rot{ScienceQA} & \rot{RealWorldQA} & \rot{Avg \tiny ± Std} &
\rot{AI2D} & \rot{Avg \tiny ± Std} &
\rot{Avg \tiny ± Std} \\
\bottomrule

\multicolumn{13}{>{\columncolor{gray!15}}l}{\textbf{Trained from scratch}} \\

HyCoCLIP-S~\cite{DBLP:conf/iclr/PalSMFGM25}
& 20.5 & 85.3
& \textbf{48.4}  & \underline{31.1}  & 43.9  & \avgstd{41.1}{7.2}
& 39.7  & \textbf{39.3}  & \avgstd{\textbf{39.5}}{0.2}
& \textbf{26.2}  & \avgstd{\textbf{26.2}}{0.0}
& \avgstd{\textbf{38.1}}{7.5} \\

MERU-S~\cite{DBLP:conf/icml/DesaiNR0V23}
& 12.0 & 85.3
& 37.5 & 29.2 & 37.6 & \avgstd{34.8}{4.8}
& 38.1 & 33.1 & \avgstd{35.6}{2.50}
& 25.5 & \avgstd{25.5}{0.0}
& \avgstd{33.5}{4.8} \\

CLIP-S~\cite{DBLP:conf/iclr/PalSMFGM25}
& 20.5 & 85.3
& 47.4  & 30.4  & 44.7  & \avgstd{40.8}{7.5}
& 39.7  & \underline{37.2}  & \avgstd{38.4}{1.3}
& 25.6  & \avgstd{25.6}{0.0}
& \avgstd{37.5}{7.6} \\
\bottomrule

\rowcolor{gray!15}
\multicolumn{13}{>{\columncolor{gray!15}}l}{\textbf{OURS 
}} \\

HAC-S w/ LN tuning
& 13.7 & \textbf{0.5}
& 47.5 & 30.6 & \underline{45.5} & \avgstd{41.2}{7.6}
& 39.3 & 35.4 & \avgstd{37.4}{2.0}
& 25.9  & \avgstd{25.9}{0.0}
& \avgstd{37.4}{7.7} \\

HAC-S w/ bias tuning
& 13.7 & \textbf{0.6}
& \underline{48.3} & 30.9 & \textbf{45.6} & \avgstd{\textbf{41.6}}{7.6}
& 39.2 & 36.5 & \avgstd{37.9}{1.3}
& \underline{26.1}  & \avgstd{\secbest{26.1}}{0.0}
& \avgstd{37.8}{7.7} \\

HAC-S w/ par. adapter
& 13.7 & \textbf{1.1}
& 47.3 & \underline{31.1} & 44.9 & \avgstd{41.1}{7.1}
& 40.1 & 36.3 & \avgstd{38.2}{1.9}
& 25.4  & \avgstd{25.4}{0.0}
& \avgstd{37.5}{7.6} \\

HAC-S w/ seq. adapter
& 13.7 & \textbf{1.1}
& 47.9 & 31.0 & 45.1 & \avgstd{\secbest{41.3}}{7.4}
& \textbf{40.4} & \underline{37.2} & \avgstd{\secbest{38.8}}{1.6}
& 25.6  & \avgstd{25.6}{0.0}
& \avgstd{\secbest{37.9}}{7.7} \\

HAC-S w/ LoRA
& 13.7 & \textbf{11.5}
& 47.5 & \textbf{31.4} & 44.6 & \avgstd{41.2}{7.0}
& 39.8 & 36.5 & \avgstd{38.2}{1.6}
& 25.5  & \avgstd{25.5}{0.0}
& \avgstd{37.6}{7.5} \\

\bottomrule
\end{tabular}
}

\label{tab:peft_search_S}
\vspace{-1.5em}
\end{table}

\subsection{Results}


Tab.~\ref{tab:peft_search_S} reports HAC-S performance across all five PEFT strategies, together with Euclidean baselines and SOTA hyperbolic models. In the General category, every method improves over the Euclidean CLIP-S baseline. For Reasoning, sequential adapters achieve the strongest result (38.8\%), with LoRA and parallel adapters both scoring 38.2\%, close to the Euclidean baseline (38.4\%). In the OCR category, all methods perform similarly, with bias tuning reaching 26.1\%. Overall, sequential adapters obtain the highest average score (37.9\%), outperforming the Euclidean baseline across all tasks despite being trained on only 13.7M samples rather than the full 20.5M used by the baseline. 
Compared to SOTA hyperbolic models, HAC-S closely approaches the HyCoCLIP-S upper bound~\cite{DBLP:conf/iclr/PalSMFGM25} (37.9\% vs. 38.1\%), despite using $85\times$ fewer trainable parameters. It also substantially outperforms MERU-S, whose average accuracy is markedly lower (33.5\%).

Tab.~\ref{tab:peft_search_B} reports HAC-B performance across all five PEFT methods. In the General category, LoRA matches the baseline average (42.0\%), followed by sequential adapters and parallel adapters achieving 41.8\% and 41.7\%, respectively. In the Reasoning category, all HAC-B variants surpass the Euclidean baseline (36.8\%), with LoRA obtaining the strongest performance (38.7\%), achieving +1.9\% average performance over the baseline. For OCR, LoRA again gives the best score (25.5\%). Notably, HAC-B equipped with LoRA matches or surpasses the Euclidean CLIP-B baseline in 5 out of 6 VQA tasks. Moreover, it outperforms the fully hyperbolic HyCoCLIP-B upperbound in four out of six tasks, despite requiring only fine-tuning and two-thirds of the training data.

\begin{table}[t]
\centering
\small
\setlength{\tabcolsep}{3pt}
\renewcommand{\arraystretch}{1.05}
\caption{Zero-shot VQA results for CLIP-B variants. 
\textbf{Bold} indicate the best result, \underline{underlined} the second-best. HyCoCLIP-B is a fully trained hyperbolic model; our HAC approaches its performance while requiring far less compute.}
\vspace{0.5em}
\resizebox{\textwidth}{!}{
\begin{tabular}{
l
C{0.90cm} C{0.90cm} 
!{\vrule width 0.6pt}
C{0.90cm} C{0.90cm} C{0.90cm} G{1.10cm} 
!{\vrule width 0.6pt}
C{0.90cm} C{0.90cm} G{1.10cm}
!{\vrule width 0.6pt}
C{0.90cm} G{1.10cm}
!{\vrule width 0.6pt}
G{1.10cm}
}

\multicolumn{1}{c}{} & \multicolumn{2}{c}{} &
\multicolumn{4}{c}{\cellcolor{gray!15}\textbf{GENERAL}} &
\multicolumn{3}{c}{\cellcolor{gray!15}\textbf{REASONING}} &
\multicolumn{2}{c}{\cellcolor{gray!15}\textbf{OCR}} &
\multicolumn{1}{c}{} \\
\cmidrule(lr){4-7}\cmidrule(lr){8-10}\cmidrule(lr){11-12}

\textbf{} &
\rot{\fontsize{8}{7.0}\selectfont train samples (M)} &
\rot{\fontsize{8}{7.0}\selectfont trainable params (M)} &
\rot{A-OKVQA} & \rot{MMStar} & \rot{SEEDBench} & \rot{Avg \tiny ± Std} &
\rot{ScienceQA} & \rot{RealWorldQA} & \rot{Avg \tiny ± Std} &
\rot{AI2D} & \rot{Avg \tiny ± Std} &
\rot{Avg \tiny ± Std} \\
\bottomrule

\multicolumn{13}{>{\columncolor{gray!15}}l}{\textbf{Trained from scratch}} \\

HyCoCLIP-B~\cite{DBLP:conf/iclr/PalSMFGM25}
& 20.5 & 149.6
& 48.6  & \underline{32.0}  & 46.3  & \avgstd{\textbf{42.3}}{7.3}
& \underline{39.2}  & 35.3  & \avgstd{37.2}{2.0}
& \textbf{26.2}  & \avgstd{\textbf{26.2}}{0.0}
& \avgstd{\secbest{37.9}}{7.8} \\

MERU-B~\cite{DBLP:conf/icml/DesaiNR0V23}
& 12.0 & 149.6
& 41.0 & 28.9 & 38.0 & \avgstd{36.0}{5.1}
& 40.4 & 36.3 & \avgstd{38.4}{2.1}
& 24.3 & \avgstd{24.3}{0.0}
& \avgstd{34.8}{6.2} \\

CLIP-B~\cite{DBLP:conf/iclr/PalSMFGM25}
& 20.5 & 149.6
& \textbf{50.0} & \textbf{31.4} & 44.7 & \avgstd{\secbest{42.0}}{7.8}
& 38.3 & 35.2 & \avgstd{36.8}{1.6}
& 25.0 & \avgstd{25.0}{0.0}
& \avgstd{37.4}{8.2} \\
\bottomrule

\rowcolor{gray!15}
\multicolumn{13}{>{\columncolor{gray!15}}l}{\textbf{OURS
}} \\

HAC-B w/ LN tuning
& 13.7 & \textbf{0.7}
& 48.8 & 31.2 & 44.5 & \avgstd{41.5}{7.5}
& 38.2 & \textbf{38.6} & \avgstd{38.4}{0.2}
& 24.8 & \avgstd{24.8}{0.0}
& \avgstd{37.7}{7.9} \\

HAC-B w/ bias tuning
& 13.7 & \textbf{0.8}
& 49.3 & 30.8 & 44.4 & \avgstd{41.5}{7.8}
& 37.9 & 37.0 & \avgstd{37.5}{0.47}
& 24.8 & \avgstd{24.8}{0.0}
& \avgstd{37.4}{8.1} \\

HAC-B w/ par. adapter
& 13.7 & \textbf{1.2}
& 49.4 & 30.7 & \textbf{45.1} & \avgstd{41.7}{8.0}
& 39.1 & 37.2 & \avgstd{38.2}{1.0}
& 25.1 & \avgstd{25.1}{0.0}
& \avgstd{37.8}{8.2} \\

HAC-B w/ seq. adapter
& 13.7 & \textbf{1.2}
& 49.4 & 31.2 & \underline{44.9} & \avgstd{41.8}{7.8}
& 39.1 & \underline{37.9} & \avgstd{\secbest{38.5}}{0.6}
& 24.6 & \avgstd{24.6}{0.0}
& \avgstd{\secbest{37.9}}{8.2} \\

HAC-B w/ LoRA
& 13.7 & \textbf{8.0}
& \underline{49.8} & \textbf{31.4} & \underline{44.9} & \avgstd{\secbest{42.0}}{7.8}
& \textbf{39.5} & \underline{37.9} & \avgstd{\textbf{38.7}}{0.8}
& \underline{25.5} & \avgstd{\secbest{25.5}}{0.0}
& \avgstd{\textbf{38.2}}{8.0} \\

\bottomrule
\end{tabular}
}

\label{tab:peft_search_B}
\vspace{-1.5em}
\end{table}

Finally, we note that best-performing adaptation strategies differ between HAC-S and HAC-B. Sequential adapters are most effective for HAC-S, while LoRA yields the best results for HAC-B. This divergence suggests that LoRA scales better with model capacity. In contrast, bias tuning drops in effectiveness from HAC-S to HAC-B, likely due to its limited complexity, as it only updates bias. Overall, these results show that more expressive PEFT methods, such as LoRA and sequential adapters, adapt differently depending on model scale. \\

\begin{figure}[th!]
\centering
\resizebox{1.0\linewidth}{!}{%
\begin{minipage}{\linewidth}
    \centering
    \begin{subfigure}{0.4\linewidth}
        \centering
        \raisebox{2.0em}{
            \includegraphics[width=\linewidth]{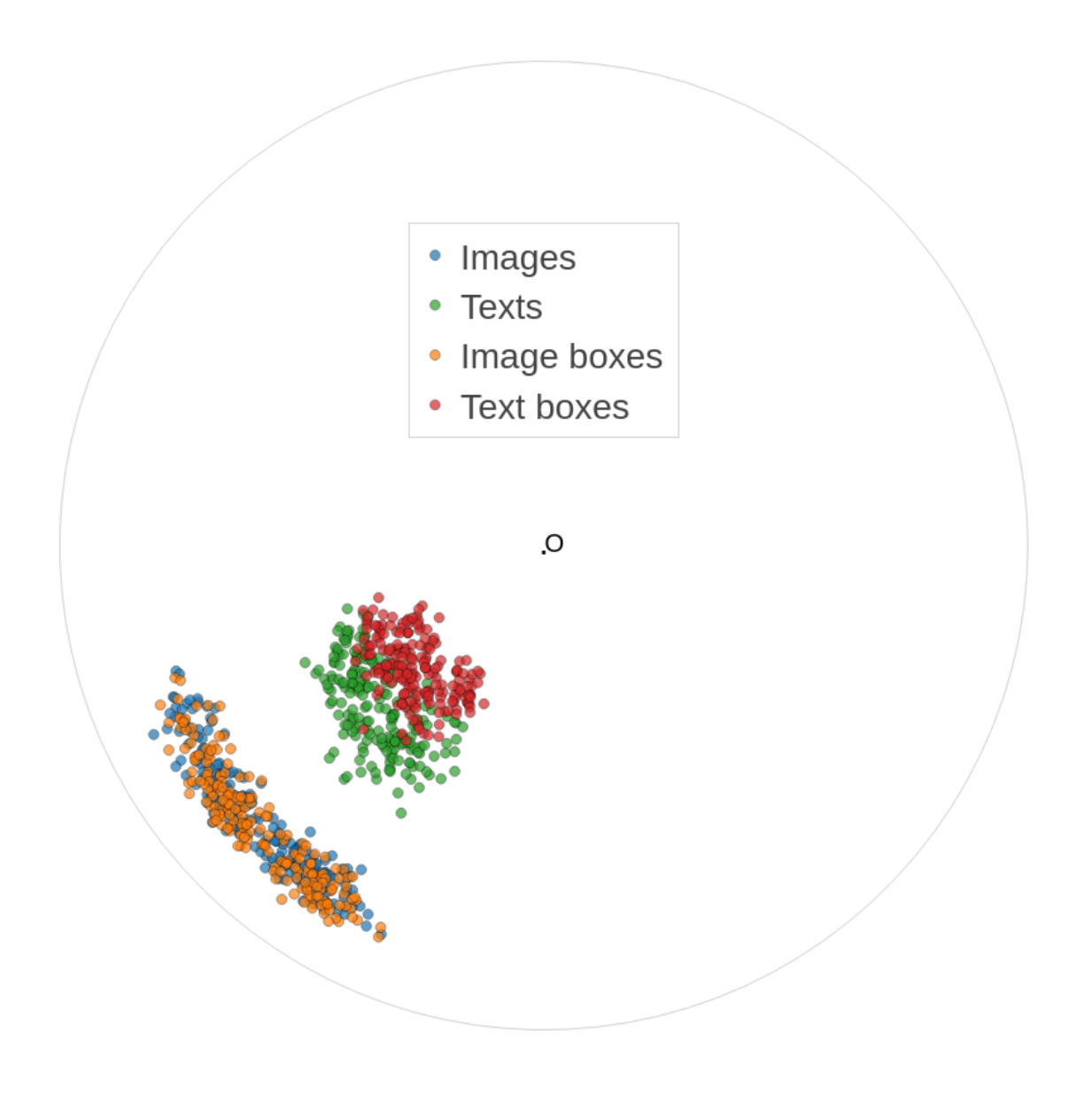}
        }
        \caption{HAC-B HoroPCA.}
    \end{subfigure}
    \hfill
    \begin{subfigure}{0.52\linewidth}
        \centering
        \includegraphics[width=\linewidth]{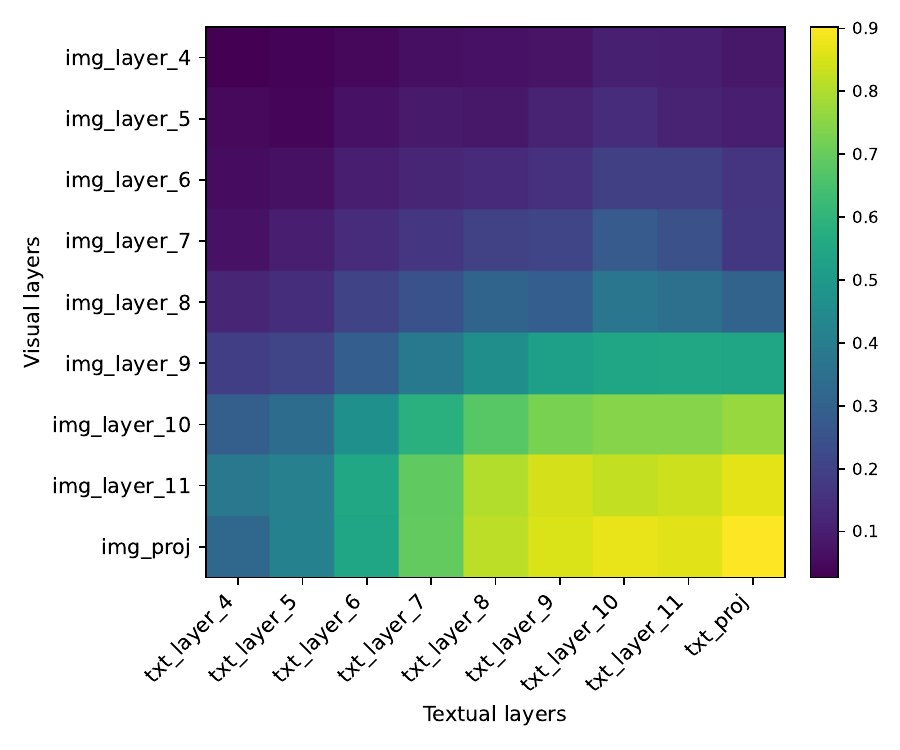}
        \caption{CLIP-B QAP alignment.}
    \end{subfigure}
\end{minipage}
}
\caption{
    (a) HoroPCA 2D projection of embeddings from our best HAC-B model.
    (b) Layer-wise QAP alignment heatmap between CLIP-B visual and textual layers, showing only layers from 4 onward to omit negligible early-layer values.}
    \label{fig:horo_vis}
    \vspace{-1.5em}
\end{figure}

\noindent \textbf{Hierarchical Geometry Learned by HAC.} 
Using embeddings from our best HAC-B model, we apply HoroPCA~\cite{DBLP:conf/icml/ChamiGNR21} to 200 GRIT image–text pairs and their object boxes. 
As shown in Fig.~\ref{fig:horo_vis}(a), text and text-box embeddings exhibit clear radial separation, with text boxes lying closest to the origin. Image and image-box embeddings, however, appear at similar radii: this pattern is also reported in~\cite{DBLP:conf/iclr/PalSMFGM25}, attributed to contrastive contraction and high visual similarity of crops to full images. Despite this, the overall radial ordering of texts, scenes and objects reflects the intended hierarchy, confirming that HAC internalizes this structure.

\vspace{-1.5em}

\subsubsection{Ablation Studies.} 
Tab.~\ref{tab:ablations} reports the contribution of each component through ablations on HAC-B.
We first vary which blocks are adapted. Holding a fixed budget of 12 trainable adapter blocks, allocating budget toward either the vision encoder (blocks 8–12) or the text encoder (blocks 4–12) lowers performance compared to our default HAC-B configuration (vision: blocks 9–12; text: blocks 5–12). This confirms that the 4-block / 8-block split used in HAC-B is the most effective.
We further support our choice of adapting more text blocks than vision blocks by performing a layer-wise alignment analysis using the Quadratic Assignment Problem (QAP) matching score~\cite{Maniparambil2024DoVA}. As shown in Fig.~\ref{fig:horo_vis}(b), the Euclidean CLIP-B text encoder becomes strongly aligned from mid-depth onward, whereas the vision encoder only reaches comparable alignment in its deepest layers. Earlier vision layers therefore act primarily as unaligned feature extractors, and adapting them harms performance. Next, we ablate the LoRA matrices: removing the attention output projection \textit{o} (\textit{q},\textit{k},\textit{v} only) or replacing attention LoRA with MLP LoRA (\textit{${fc}_{1}$},\textit{${fc}_{2}$} only), leads to a drop in VQA accuracy, indicating that the full attention set (\textit{q},\textit{k},\textit{v},\textit{o}) is important. Reducing instead the LoRA rank progressively harms performance, showing that geometric adaptation of the CLIP-B model benefits from high-rank updates. 
Notably, when we apply the same strategy to CLIP-S, increasing either the number of trainable submatrices or the rank, does not yield any improvement (see Supplementary Material for details). This indicates that optimal hyperparameters depend on model scale, a trend also observed in prior work investigating Transformers, e.g.,~\cite{DBLP:conf/emnlp/IvgiCB22}.

We further investigate the effect of removing the Entailment Loss $\mathcal{L}_{\mathrm{hCE}}$ from our final loss function. Training with only contrastive loss $\mathcal{L}_{\mathrm{hCC}}$ leads to a consistent accuracy degradation across HAC-S and HAC-B, showing that the absence of $\mathcal{L}_{\mathrm{hCE}}$ collapses the learned hyperbolic geometry. Without this loss, the model can no longer maintain the object-scene partial order that hyperbolic geometry is intended to encode, and the resulting embeddings lose their radial organization. The corresponding numeric results, together with the associated qualitative geometric structure learned (i.e., hierarchies) with or without this loss, are given in the Supplementary Material.

\begin{table}[t]
\centering
\setlength{\tabcolsep}{6pt}
\caption{Ablations of HAC-B on CLIP blocks, LoRA ranks, and LoRA matrices.}
\resizebox{1.0\linewidth}{!}{%
\begin{tabularx}{\linewidth}{l *{5}{>{\centering\arraybackslash}X}}
\cmidrule(lr){2-6}
{} & Vision blocks & Text blocks & LoRA matrices & LoRA rank & \raisebox{-1.3ex}{Mean(6)} \\
\bottomrule
\rowcolor{gray!15}
LoRA rank & & & & & \\
{} & 9-12  & 5-12  & q,k,v,o & 8 & 37.4 \\
{} & 9-12  & 5-12  & q,k,v,o & 32 & 37.9 \\
\rowcolor{gray!15}
LoRA matrices & & & & & \\
{} & 9-12  & 5-12  & q,k,v & 128 & 38.0 \\
{} & 9-12  & 5-12  & \(\mathrm{fc}_1,\,\mathrm{fc}_2\) & 128 & 38.0 \\
\rowcolor{gray!15}
CLIP blocks & & & & & \\
{} & 8-12 & 6-12 & q,k,v,o & 128 & 38.1 \\
{} & 10-12 & 4-12 & q,k,v,o & 128 & 38.1 \\
\midrule
\textbf{HAC-B} & \textbf{9-12}  & \textbf{5-12}  & \textbf{q,k,v,o} & \textbf{128} & \textbf{38.2} \\
\bottomrule
\end{tabularx}
}

\label{tab:ablations}
\vspace{-1.5em}
\end{table}



\section{Conclusions}
\vspace{-0.5em}
In this work, we introduced HAC, a parameter-efficient framework for adapting pretrained Euclidean CLIP models to hyperbolic space for zero-shot VQA. By leveraging lightweight adaptation modules, we showed that HAC can reshape CLIP's embedding geometry without full-model retraining, resulting in consistent improvements over Euclidean baselines, and even surpassing fully hyperbolic models with reduced parameters and data budgets. While HAC shows empirical gains, it also has limitations. 
In fact, while hyperbolic geometry offers a natural hierarchical structure, it may not capture all types of reasoning needed for VQA. Future work will explore 
HAC application to multimodal tasks beyond VQA, such as zero-shot image classification.

\subsubsection{\ackname} We acknowledge ISCRA for awarding this project access to the LEONARDO supercomputer, owned by the EuroHPC Joint Undertaking, hosted by CINECA (Italy).

\section{Supplementary Material}

\subsection{Further Implementation Details}
We employ a set of standard image augmentations during training; below, we list all transformations together with their associated hyperparameters and probabilities:

\begin{itemize}
    \item Resize to 224 on the shorter side and center crop (always applied)
    \item Random horizontal flip ($p = 0.5$)
    \item Color jitter: brightness = 0.2, contrast = 0.2, saturation = 0.2, hue = 0.05 ($p = 0.5$)
    \item Gaussian blur: kernel size = 3, $\sigma \sim \mathcal{U}(0.1, 0.5)$ ($p = 0.2$)
    \item AutoContrast ($p = 0.1$)
    \item Histogram equalization ($p = 0.05$)
    \item Sharpness adjustment: factor = 1.1 ($p = 0.1$)
    \item Gamma correction: $\gamma \sim \mathcal{U}(0.9, 1.1)$ ($p = 0.2$)
\end{itemize}

\subsection{Further Ablation Studies}
\subsubsection{Ablation on LoRA Matrices and Rank for HAC-S}
We analyze the effect of increasing the number of LoRA matrices and the LoRA rank when adapting our HAC-S model. In contrast to HAC-B, expanding LoRA to additional attention submatrices, or raising the rank beyond 8, does not yield further improvements and slightly degrades performance under the HAC-S configuration. A summary of these observations is provided in Table~\ref{tab:lora_supp}.

\begin{table}[ht]
\centering
\caption{Ablation study of HAC-S on LoRA ranks and LoRA matrices.}
\label{tab:lora_supp}

\setlength{\tabcolsep}{6pt}
\resizebox{1.0\linewidth}{!}{%
\begin{tabularx}{\linewidth}{l *{5}{>{\centering\arraybackslash}X}}
\cmidrule(lr){2-4}
{} & LoRA matrices & LoRA rank & \raisebox{0ex}{Mean(6)} \\
\bottomrule
\rowcolor{gray!15}
LoRA rank & & & \\
{} & q,v & 32 & 37.0 \\
{} & q,v & 16 & 37.2 \\
\rowcolor{gray!15}
LoRA matrices & & & \\
{} & q,k,v & 8 & 37.3 \\
{} & q,k,v,o & 8 & 37.4 \\
\midrule
\textbf{HAC-S} & \textbf{q,v} & \textbf{8} & \textbf{37.6} \\
\bottomrule
\end{tabularx}
}
\end{table}

\subsubsection{Ablation on Entailment Loss}
To better understand the role of the hierarchical Compositional Entailment loss $\mathcal{L}_{\mathrm{hCE}}$ in structuring the hyperbolic embedding space, we perform an ablation in which the model is trained using only the contrastive component $\mathcal{L}_{\mathrm{hCC}}$. 

As shown in Table~\ref{tab:loss_ablation}, removing $\mathcal{L}_{\mathrm{hCE}}$ consistently degrades VQA performance for both HAC-S and HAC-B, with the effect being substantially more pronounced in the larger HAC-B model ($-0.7$ vs. $-0.1$ mean accuracy). This confirms that $\mathcal{L}_{\mathrm{hCE}}$ is key to enforcing the hierarchical partial order that hyperbolic geometry is designed to represent. 

\begin{figure}[ht!]
    \centering

    \begin{subfigure}[t]{0.45\textwidth}
        \centering
        \includegraphics[width=\linewidth]{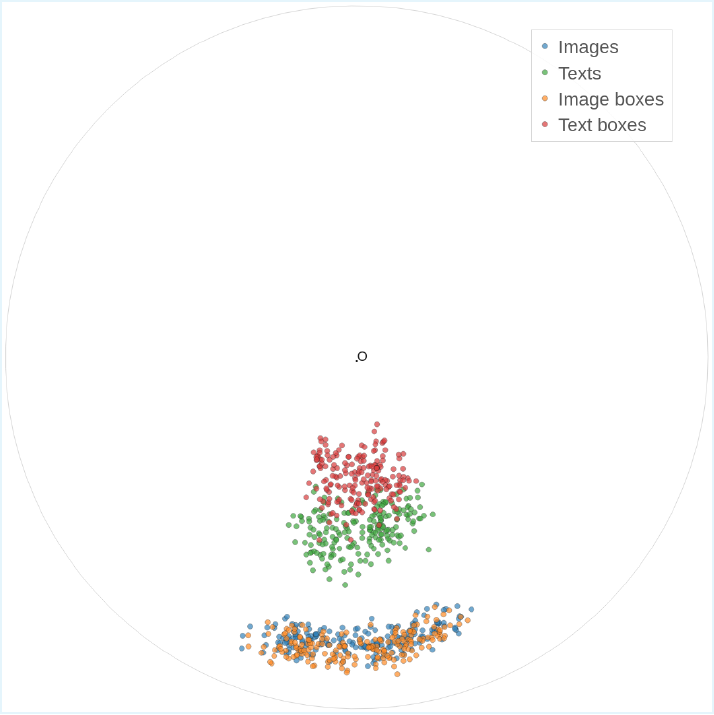}
        \caption{HAC-S w/ $\mathcal{L}_{\mathrm{hCE}}$}
    \end{subfigure}
    \hfill
    \begin{subfigure}[t]{0.45\textwidth}
        \centering
        \includegraphics[width=\linewidth]{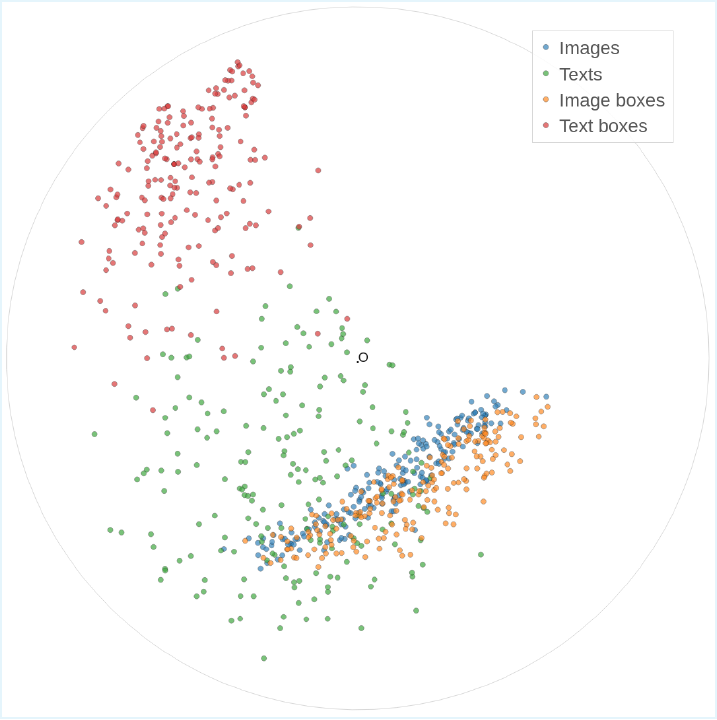}
        \caption{HAC-S w/o $\mathcal{L}_{\mathrm{hCE}}$}
    \end{subfigure}
    \vspace{15pt}

    \begin{subfigure}[t]{0.45\textwidth}
        \centering
        \includegraphics[width=\linewidth]{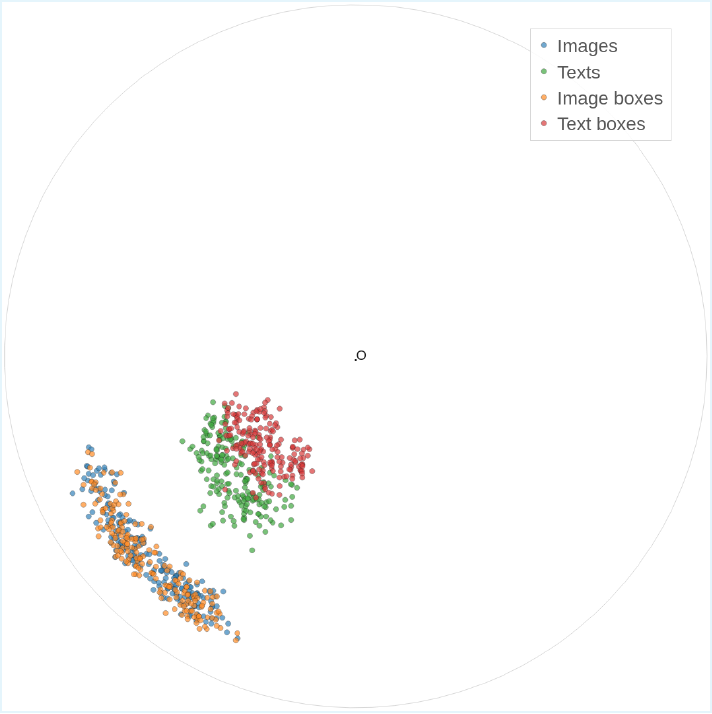}
        \caption{HAC-B w/ $\mathcal{L}_{\mathrm{hCE}}$}
    \end{subfigure}
    \hfill
    \begin{subfigure}[t]{0.45\textwidth}
        \centering
        \includegraphics[width=\linewidth]{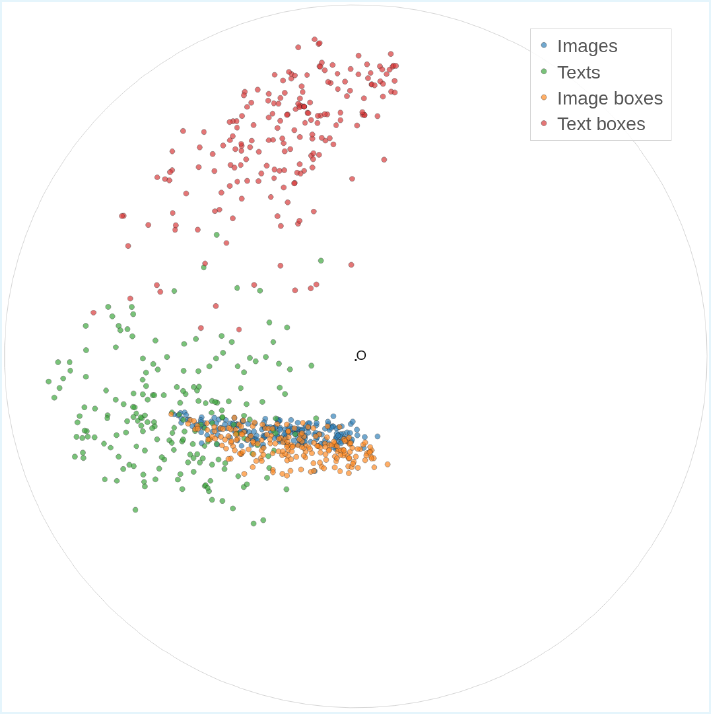}
        \caption{HAC-B w/o $\mathcal{L}_{\mathrm{hCE}}$}
    \end{subfigure}

    \caption{\textbf{HoroPCA projections} comparing the geometric structure learned with and without Compositional Entailment Loss $\mathcal{L}_{\mathrm{hCE}}$:\newline
    \textbf{(a) HAC-S with $\bm{\mathcal{L}}_{\bm{\mathrm{hCE}}}$}: clear hierarchical separation, with object boxes near the origin and scene-level embeddings at larger radii.\newline
    \textbf{(b) HAC-S without $\bm{\mathcal{L}}_{\bm{\mathrm{hCE}}}$}: loss of hierarchical organization; embeddings lack radial ordering.\newline
    \textbf{(c) HAC-B with $\bm{\mathcal{L}}_{\bm{\mathrm{hCE}}}$}: clear object-scene hierarchy.\newline
    \textbf{(d) HAC-S without $\bm{\mathcal{L}}_{\bm{\mathrm{hCE}}}$}: hierarchy collapses; embeddings disperse without respect to partial order.}
    \label{fig:entail_ablation}
\end{figure}

The qualitative HoroPCA projections in Fig.~\ref{fig:entail_ablation} further illustrate this phenomenon. When $\mathcal{L}_{\mathrm{hCE}}$ is active, both HAC-S and HAC-B exhibit a clear radial organization: object-box embeddings cluster closer to the origin, while scene-level image embeddings occupy outer radii, reflecting the intended object-scene hierarchy. However, when $\mathcal{L}_{\mathrm{hCE}}$ is omitted, this structure collapses: embeddings become distributed without any coherent ordering from the root, and objects no longer occupy the more central regions of the manifold. Overall, these results demonstrate that $\mathcal{L}_{\mathrm{hCE}}$ is essential for preserving hyperbolic hierarchy and for unlocking the full benefits of HAC, especially in larger models.

\begin{table}[ht]
\centering
\caption{Ablation study of Entailment Loss $\mathcal{L}_{\mathrm{hCE}}$ for HAC-S and HAC-B.}
\label{tab:loss_ablation}

\setlength{\tabcolsep}{6pt}
\resizebox{1.0\linewidth}{!}{%
\begin{tabularx}{\linewidth}{l *{3}{>{\centering\arraybackslash}X}}
\cmidrule(lr){2-4}
{} & $\mathcal{L}_{\mathrm{hCC}}$ & $\mathcal{L}_{\mathrm{hCE}}$ & \raisebox{0ex}{Mean(6)} \\
\bottomrule
\rowcolor{gray!15}
HAC-S w/ seq. adapters & & & \\
{} & \cmark & \xmark & 37.8 \\[0.3em]
{} & \cmark & \cmark & \textbf{37.9} \\[0.3em]
\rowcolor{gray!15}
\bottomrule
HAC-B w/ LoRA & & & \\
{} & \cmark & \xmark & 37.5 \\[0.3em]
{} & \cmark & \cmark & \textbf{38.2} \\[0.3em]
\bottomrule
\end{tabularx}
}
\end{table}

%
%
\bibliographystyle{splncs04}
\bibliography{bibliography}

\end{document}